\documentclass{article}

\usepackage{microtype}
\usepackage{graphicx}
\usepackage{subcaption}
\usepackage{booktabs} % for professional tables
\PassOptionsToPackage{hyphens}{url}
\usepackage{hyperref}
\usepackage{longtable}
\usepackage{listings}
\usepackage{amsmath}
\usepackage{amssymb}
\usepackage{mathtools}
\usepackage{amsthm}
\usepackage{multirow}
\usepackage{xcolor}
\usepackage{bm}

\usepackage{caption} 
\usepackage{graphicx}

\usepackage[preprint]{icml2026}

\theoremstyle{plain}

\theoremstyle{definition}

\theoremstyle{remark}

\colorlet{punct}{red!60!black}
\definecolor{background}{HTML}{EEEEEE}
\definecolor{delim}{RGB}{20,105,176}
\colorlet{numb}{magenta!60!black}

\lstdefinelanguage{json}{
    basicstyle=\normalfont\ttfamily\small,
    numbers=left,
    numberstyle=\scriptsize,
    stepnumber=1,
    numbersep=8pt,
    showstringspaces=false,
    breaklines=true,
    frame=lines,
    literate=
     *{0}{{{\color{numb}0}}}{1}
      {1}{{{\color{numb}1}}}{1}
      {2}{{{\color{numb}2}}}{1}
      {3}{{{\color{numb}3}}}{1}
      {4}{{{\color{numb}4}}}{1}
      {5}{{{\color{numb}5}}}{1}
      {6}{{{\color{numb}6}}}{1}
      {7}{{{\color{numb}7}}}{1}
      {8}{{{\color{numb}8}}}{1}
      {9}{{{\color{numb}9}}}{1}
      {:}{{{\color{punct}{:}}}}{1}
      {,}{{{\color{punct}{,}}}}{1}
      {\{}{{{\color{delim}{\{}}}}{1}
      {\}}{{{\color{delim}{\}}}}}{1}
      {[}{{{\color{delim}{[}}}}{1}
      {]}{{{\color{delim}{]}}}}{1},
}

\usepackage[textsize=tiny]{todonotes}

\icmltitlerunning{AIR-VLA: Vision-Language-Action Systems for Aerial Manipulation}

\begin{document}

\twocolumn[
  \icmltitle{AIR-VLA: Vision-Language-Action Systems for Aerial Manipulation}

  \icmlsetsymbol{corr}{*}

  \begin{icmlauthorlist}
    \icmlauthor{Jianli Sun}{casia}
    \icmlauthor{Bin Tian}{casia}
    \icmlauthor{Qiyao Zhang}{bit}
    \icmlauthor{Chengxiang Li}{syu}
    \icmlauthor{Zihan Song}{hnu}
    \icmlauthor{Zhiyong Cui}{bhu}
    \icmlauthor{Yisheng Lv}{casia}
    \icmlauthor{Yonglin Tian}{casia,corr}
  \end{icmlauthorlist}

  \icmlaffiliation{casia}{The Institute of Automation, Chinese Academy of Sciences, Beijing 100190, China}
  \icmlaffiliation{bit}{School of Automation, Beijing Institute of Technology, Beijing 100081, China}
  \icmlaffiliation{syu}{School of Information and Intelligent Engineering, University of Sanya, Sanya 572000, Hainan Province, China}
  \icmlaffiliation{hnu}{School of Mechanical and Vehicle Engineering, Hunan University, Changsha 410082, Hunan Province, China}
  \icmlaffiliation{bhu}{State Key Lab of Intelligent Transportation Systems, School of Transportation Science and Engineering, Beihang University, Beijing, China}

  \icmlcorrespondingauthor{Yonglin Tian}{yonglin.tian@ia.ac.cn}

  \icmlkeywords{Machine Learning, ICML, Aerial Manipulation, Vision-Language-Action Models}

  \vskip 0.3in

  % =================================================================
  % 【核心修改】这里开始插入第一页大图
  % 原理：不使用 figure 环境，直接插入图片，用 captionof 生成编号
  % =================================================================
  {
    \centering  % 使图片居中
    \includegraphics[width=0.95\textwidth]{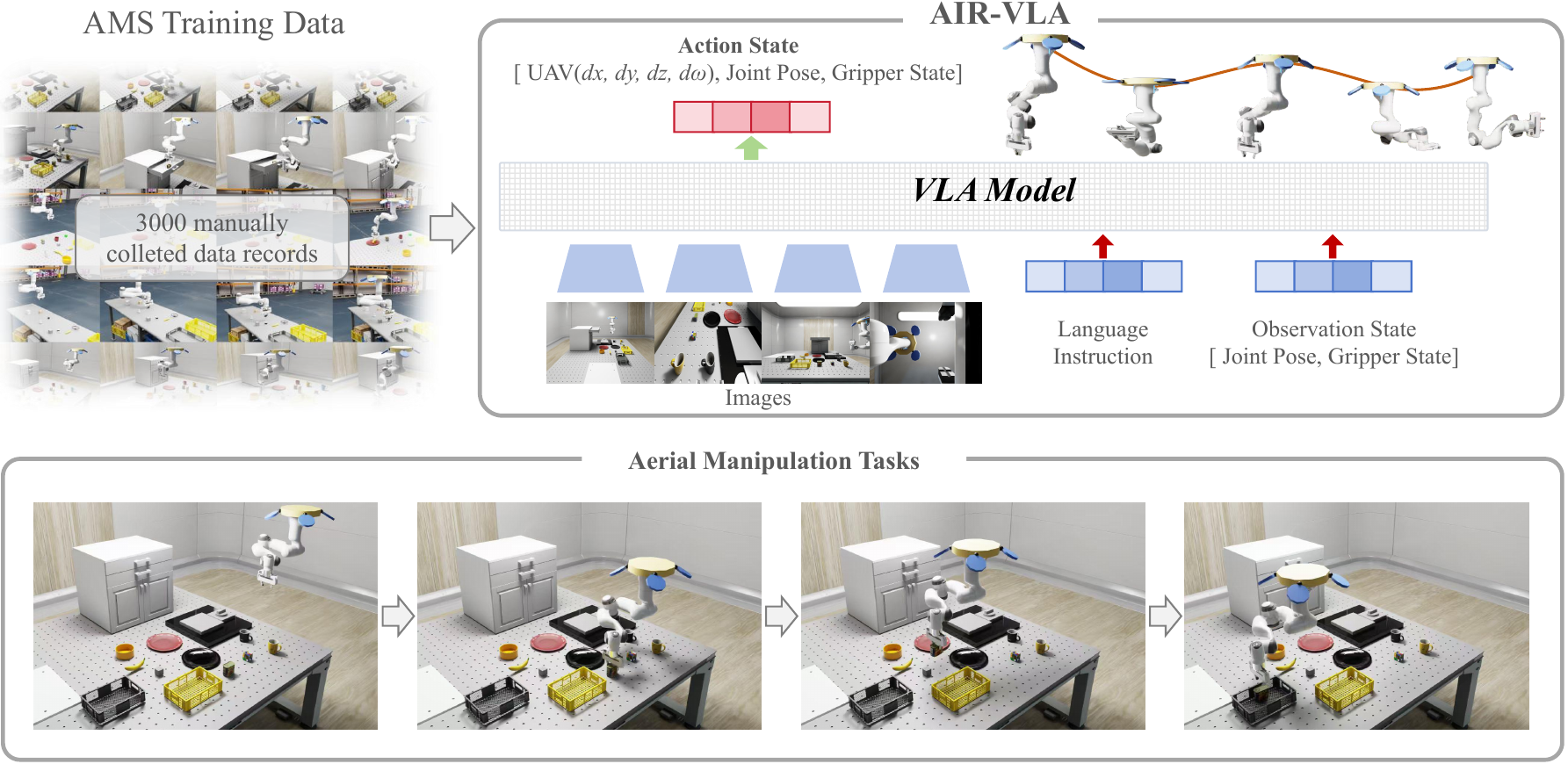} % 建议宽度设为 0.95 左右，留点白边好看
    \par        % 强制换行（关键！防止报错）
    \vspace{10pt} % 图片和标题之间的距离
    
    % 手动生成 Figure 1 的标题
    \captionof{figure}{\textbf{Overview of AIR-VLA model framework and dataset of aerial manipulation tasks.}}
    \label{fig:air_vla_overview} % 给它一个标签，正文可以用 \ref 引用
    \par        % 再次强制换行
  }
  % =================================================================
  
  \vskip 0.2in

]

\printAffiliationsAndNotice{}

\begin{abstract}
While Vision-Language-Action (VLA) models have achieved remarkable success in ground-based embodied intelligence, their application to Aerial Manipulation Systems (AMS) remains a largely unexplored frontier. The inherent characteristics of AMS, including floating-base dynamics, strong coupling between the UAV and the manipulator, and the multi-step, long-horizon nature of operational tasks, pose severe challenges to existing VLA paradigms designed for static or 2D mobile bases. To bridge this gap, we propose \textbf{AIR-VLA}, the first VLA benchmark specifically tailored for aerial manipulation. We construct a physics-based simulation environment and release a high-quality multimodal dataset comprising 3000 manually teleoperated demonstrations, covering base manipulation, object \& spatial understanding, semantic reasoning, and long-horizon planning. Leveraging this platform, we systematically evaluate mainstream VLA models and state-of-the-art VLM models. Our experiments not only validate the feasibility of transferring VLA paradigms to aerial systems but also, through multi-dimensional metrics tailored to aerial tasks, reveal the capabilities and boundaries of current models regarding UAV mobility, manipulator control, and high-level planning. \textbf{AIR-VLA} establishes a standardized testbed and data foundation for future research in general-purpose aerial robotics. The resource of AIR-VLA will be available at https://github.com/SpencerSon2001/AIR-VLA.
\end{abstract}

\section{Introduction}

With breakthrough advancements in Large Language Models (LLMs) and Vision-Language Models (VLMs), Embodied AI is undergoing a paradigm shift from specialized skill learning to general-purpose intelligence. Recently, Vision-Language-Action (VLA) models, represented by RT-1 \cite{brohan2023rt1roboticstransformerrealworld}, OpenVLA \cite{kim2024openvlaopensourcevisionlanguageactionmodel}, and $\pi_0$ \cite{black2026pi0visionlanguageactionflowmodel}, have demonstrated exceptional capability in handling open-world tasks driven by natural language instructions. However, existing VLA research is predominantly confined to Ground Mobile Manipulators, where the operational space is restricted to 2D planar navigation and limited working heights. This significantly constrains the potential of embodied intelligence in exploring broader three-dimensional physical spaces.

As a pivotal yet challenging branch of embodied intelligence, Aerial Manipulation Systems (AMS) integrate the high mobility of multi-rotor UAVs with the dexterous manipulation capabilities of robotic arms. By transcending the physical limitations of terrain and altitude, AMS demonstrate irreplaceable value in complex scenarios such as logistics transportation, disaster rescue, infrastructure inspection, and high-altitude operations. However, extending VLA models to aerial platforms introduces unique physical and control challenges. AMS must contend not only with the instability of a floating base and the kinematic coupling between the UAV and the manipulator, but also with the demands of 3D spatial reasoning within an expansive workspace and the inherently long-horizon, multi-step nature of operational tasks. Consequently, this requires that models possess not only general semantic understanding, but also the mastery of complex coordination strategies, ranging from 3D omnidirectional navigation to fine-grained end-effector manipulation, along with high-level planning capabilities for complex missions.

Despite its promising prospects, research on VLA for aerial manipulation currently faces a severe ``data scarcity'' and an ``absence of benchmarks''. Existing embodied intelligence benchmarks are predominantly designed for tabletop manipulation\cite{NEURIPS2023_8c3c6668} or ground navigation\cite{yenamandra2023homerobot}. Consequently, they fail to effectively evaluate omnidirectional mobility in 3D space, manipulation robustness under floating-base conditions, and long-horizon temporal reasoning capabilities conditioned on language. Furthermore, current research applying VLA to the UAV domain remains focused on navigation\cite{lykov2025cognitivedronevlamodelevaluation}, task generation\cite{sautenkov2025uavvlavisionlanguageactionlargescale}, or single-DoF VLA manipulation tightly coupled with path planning\cite{mehboob2026dronevlavlabasedaerial}, leaving the application of VLA to full-spectrum aerial manipulation tasks largely unexplored.

However, with the maturation of AMS hardware and control theory, combined with the rapid advancement of VLMs in robotics, applying the scene generalization capabilities of VLAs and the high-level planning abilities of VLMs to aerial manipulation tasks has demonstrated immense potential. Therefore, there is an urgent need in the academic community for a dataset and evaluation framework specifically focused on aerial manipulation.

To address these challenges, we propose \textbf{AIR-VLA}, the first VLA training and evaluation benchmark designed specifically for Aerial Manipulation Systems. We construct a simulation environment integrated with a physics engine based on NVIDIA Isaac Sim, simulating the collaborative operation of UAVs and manipulators under complex dynamic conditions. On this basis, we construct and release a multimodal dataset covering tasks ranging from basic primitive actions to complex 3D dynamic interactions requiring long-horizon planning. Spanning diverse simulated scenarios, covering indoor rooms, factories, and outdoor environments, the dataset comprises \textbf{3000 high-quality manually teleoperated aerial manipulation episodes}. \textbf{AIR-VLA} aims to provide a comprehensive performance profile of aerial embodied intelligence across multiple dimensions, including basic motion control, fine-grained visual and spatial understanding, semantic instruction following, and long-horizon task planning.

The main contributions of this paper are summarized as follows:

\begin{itemize}
    \item \textbf{Pioneering Aerial Manipulation VLA Benchmark:} We propose the first VLA benchmark testbed specifically designed for AMS, filling the evaluation gap in the domain of 3D aerial manipulation and providing a standardized research tool for the community. Tailored to the unique characteristics of aerial operations, we design a multi-suite dataset rich in sensory information (RGB, depth, proprioception) and diverse language instructions, providing high-quality data support for training aerial VLA manipulation policies.

    \item \textbf{Tailored Evaluation Metrics:} Addressing the unique characteristics of aerial manipulation, specifically the tight coordination between the UAV and manipulator and the long-horizon, multi-step nature of tasks, we move beyond relying solely on task success rates. We introduce multi-dimensional evaluation metrics designed to assess the aerial manipulation performance of VLAs and the long-horizon planning capabilities of VLMs. These metrics, which cover dimensions such as UAV localization, navigation, and spatial perception, allow a deep exploration of the performance boundaries of the models.

    \item \textbf{Comprehensive Baseline Analysis:} We conduct extensive benchmarking and in-depth analysis of mainstream VLA and VLM models on \textbf{AIR-VLA}. By quantifying the performance of current mainstream VLA models on aerial tasks and the high-level planning capabilities of VLMs, we reveal critical challenges in the transfer process from ground to aerial platforms and provide guidance for future research directions.
\end{itemize}

\begin{figure*}[t] 
  \vskip 0.2in
  \begin{center}
    \centerline{\includegraphics[width=\textwidth]{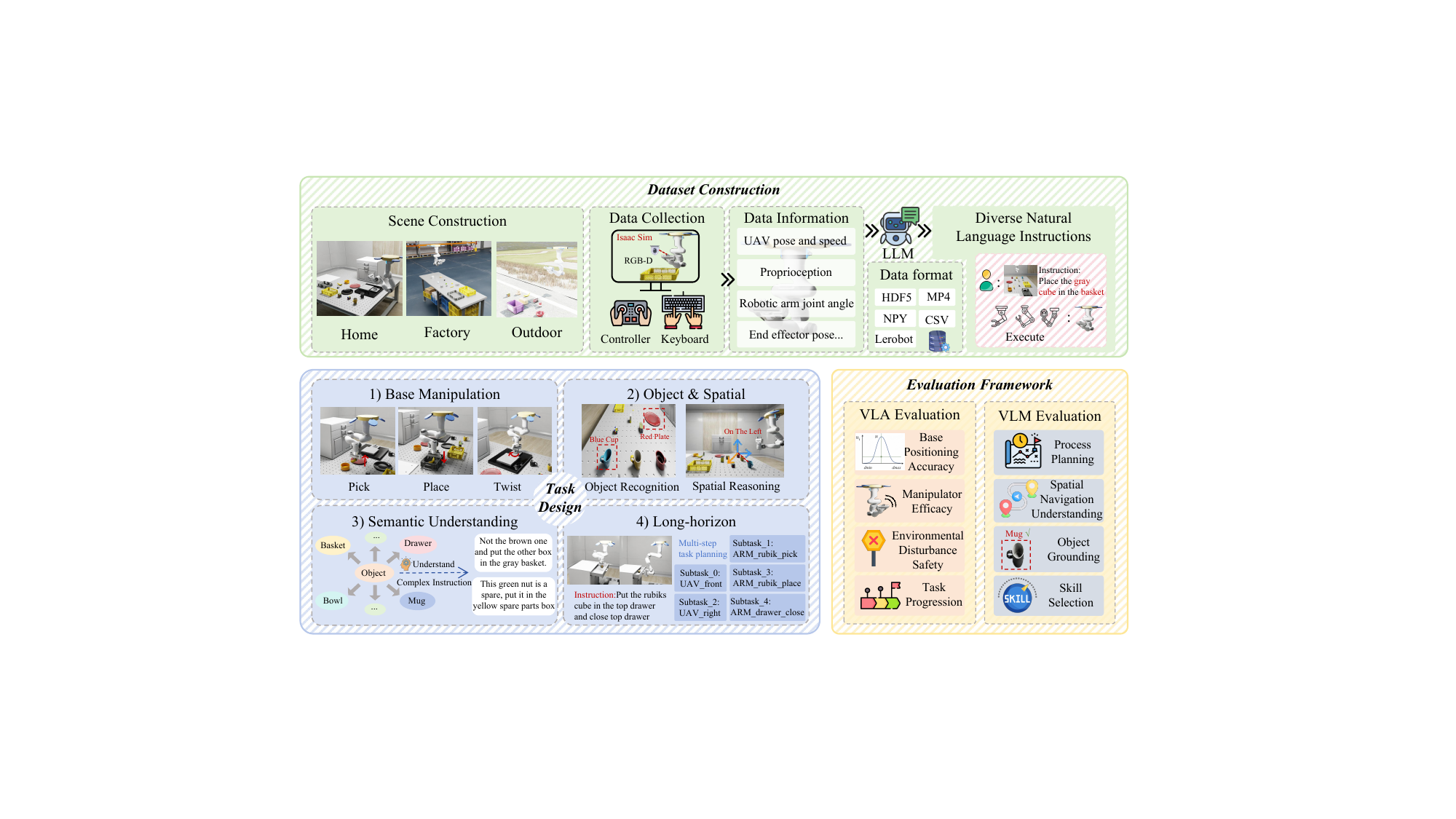}}
    \caption{
      \textbf{Overview of the AIR-VLA benchmark.} AIR-VLA serves as a full-stack Vision-Language-Action platform tailored for aerial manipulation systems. It integrates a simulation-based teleoperation data acquisition pipeline, an online simulation environment, and diverse multimodal datasets. Furthermore, it provides a comprehensive benchmark for evaluating mainstream VLA and VLM models across varied aerial manipulation tasks.
    }
    \label{fig:air_vla_overview} 
  \end{center}
\end{figure*}

\section{Related Works}

\subsection{Aerial Manipulation Systems}
AMS extend UAV capabilities from passive reconnaissance to active physical interaction, though the coupling between the floating base and manipulator introduces severe nonlinear dynamics. 
In mechanical design, research has prioritized bio-inspired and compliant architectures to enhance adaptability and stability, such as owl-inspired limbs \cite{Liu2024} and continuum elephant-trunk manipulators \cite{Peng2025}. Dual-arm systems have also been explored to execute complex multi-step tasks infeasible for single-arm platforms \cite{Ghorbani2023}. 
On the control frontier, strategies address external disturbances and uncertainties through refined anti-disturbance techniques \cite{Liu2024}, adaptive prescribed performance control \cite{9812607}, and Reinforcement Learning (RL)-driven hierarchical policies \cite{deshmukh2025globalendeffectorposecontrol}. Additionally, visual impedance control \cite{9750110} and whole-body planning \cite{10547187} are employed to improve interaction precision. 
Despite these advances, traditional AMS frameworks generally lack semantic grounding. Operating under closed-world assumptions prevents them from generalizing to unstructured environments or interpreting abstract instructions \cite{10943237, carvajal2024multitask, 11219352}, limiting their utility as intelligent assistants.

\subsection{Vision-Language-Action Models}
VLA models shift Embodied AI from modular pipelines to end-to-end architectures, unifying perception, language, and action into a single transformer backbone. Foundational models like RT-1 \cite{brohan2023rt1roboticstransformerrealworld} and OpenVLA \cite{kim2024openvlaopensourcevisionlanguageactionmodel} demonstrate that large-scale training enables open-vocabulary task execution. 
The field is rapidly evolving toward high-frequency control and long-horizon capabilities. The $\pi$ series, including $\pi_0$ \cite{black2026pi0visionlanguageactionflowmodel} and $\pi_{0.5}$ \cite{black2025pi}, introduces flow-matching for smooth action generation. Notably, $\pi_{0.6}$ \cite{intelligence2025pi06vlalearnsexperience} exhibits robust autonomous execution of long-horizon tasks, indicating that VLAs are maturing into reliable systems for complex physical interaction.

\subsection{VLA for Aerial Robotics}
Recognizing VLA potential, recent works have integrated these models into aerial platforms, focusing primarily on navigation and trajectory generation. 
Benchmarks like \textit{IndoorUAV} \cite{liu2025indooruavbenchmarkingvisionlanguageuav} leverage task decomposition and $\pi_0$ for 3D language grounding, while \textit{RaceVLA} \cite{serpiva2025racevlavlabasedracingdrone} enables high-speed agility via video-conditioned control. To address onboard constraints, frameworks utilizing 3D Gaussian Splatting \cite{wu2025vlaanefficientonboardvisionlanguageaction}, fine-grained instruction mapping \cite{wang2025uavflowcolosseorealworldbenchmark}, and cognitive reasoning fusion \cite{lykov2025cognitivedronevlamodelevaluation} have been proposed.
However, current applications remain predominantly confined to trajectory planning or high-level mission generation. Approaches like UAV-VLA \cite{sautenkov2025uavvlavisionlanguageactionlargescale} and DroneVLA \cite{mehboob2026dronevlavlabasedaerial} largely limit the VLA's role to mission planning or simple 1-DoF gripping. These methods fail to leverage the UAV's potential for complex physical interaction in unstructured environments. The lack of a mature framework for full-spectrum aerial manipulation represents a critical research gap.

\subsection{Benchmarks for Robot Manipulation}
High-quality benchmarks drive VLA research but currently focus on ground-based platforms. 
For static manipulation, \textit{LIBERO} \cite{NEURIPS2023_8c3c6668} and \textit{RoboTwin 2.0} \cite{chen2025robotwin20scalabledata} evaluate lifelong learning and dual-arm coordination, respectively. For mobile manipulation, \textit{OVMM} \cite{yenamandra2023homerobot} and \textit{Mobile ALOHA} \cite{fu2024mobile} address open-vocabulary tasks and low-cost imitation learning. 
While comprehensive for ground robots, these benchmarks are ill-suited for AMS. Specifically, they lack the robotic platforms, corresponding kinematic designs, and aerial manipulation task data specific to Aerial Manipulation Systems. This absence restricts the development of aerial VLA models. To bridge this gap, we propose \textit{AIR-VLA}, the first benchmark specifically designed to extend VLA intelligence to the challenging realm of aerial mobile manipulation.

\section{AIR-VLA}

To comprehensively evaluate the generalization capability and policy robustness of VLA models in the aerial manipulation domain, AIR-VLA is systematically designed across four dimensions: task taxonomy, evaluation framework, simulation environment, and dataset construction. This design aims to provide standardized metrics while simulating authentic physical characteristics.

\subsection{Task Design Taxonomy}
AIR-VLA comprises four core task suites targeting manipulation precision, semantic reasoning, visual perception, and planning. Unlike ground-based benchmarks, it emphasizes deep coordination between 3D UAV positioning and manipulator actions, covering a spectrum from atomic actions to complex interactions. Tailored to AMS characteristics, the dataset features an average task length of approximately 475 time steps, significantly exceeding traditional benchmarks to reflect temporal complexity.The specific objectives of the four suites are:

\begin{itemize}
    \item \textbf{Base Manipulation:} Evaluates low-level motion control and UAV-arm coordination with minimal distractors. Tasks prioritize optimal 3D hovering and precise manipulation, requiring the model to handle complex contact dynamics.
     
    \item \textbf{Object \& Spatial:} Assesses fine-grained understanding of physical attributes (color, shape, material) and spatial relationships. This suite imposes stringent requirements on cross-modal alignment and 3D geometric reasoning.
     
    \item \textbf{Semantic Understanding:} Tests robustness to unstructured language instructions. Tasks feature diverse styles and implicit intents, requiring the model to infer underlying goals from complex natural language.
     
    \item \textbf{Long-Horizon:} Evaluates multi-step reasoning with temporal dependencies. The AMS must alternate between wide-range maneuvers and local fine-grained operations. Any coordination failure in the chain results in total task failure, representing the most challenging scenario.
\end{itemize}

\subsection{Evaluation Framework}

Compared to traditional ground robot tasks, aerial mobile manipulation introduces unique challenges such as dynamic coupling of the floating base, volumetric workspaces, and temporal complexity of long-horizon tasks. To comprehensively evaluate the limit performance of embodied intelligence models in this high-dimensional domain, we constructed a two-layer evaluation framework covering perception, planning, and control. For VLA models, we conduct closed-loop evaluations based on an online simulation environment, focusing on UAV-Arm Coordination capabilities under real-time inference. Specifically, we quantify the feasibility of UAV spatial navigation, the manipulation precision of the end-effector, and dynamic safety during long-horizon task execution. For long-horizon tasks, considering the critical role of logical reasoning, we propose a multi-dimensional evaluation metric set for the high-level task planning capabilities of VLMs. This assesses the embodied reasoning ability of VLMs from dimensions including sub-task process construction, 3D spatial perception, fine-grained object grounding, and atomic skill selection. \textbf{For detailed definitions and mathematical formulations of these metrics, please refer to Appendix A.3.}

\subsection{Simulation Environment}
\textbf{Simulator.} Built on NVIDIA Isaac Sim, AIR-VLA leverages the Omniverse platform to bridge the Sim-to-Real gap. We utilize the PhysX 5 engine and ray-tracing to ensure realistic simulation from fluid dynamic disturbances to lighting reflections. Furthermore, the platform's USD pipeline and GPU parallel acceleration enable flexible scene configuration and efficient data collection.

\textbf{Robot System.} To fully test the adaptability of VLA models to high-dimensional state spaces, we constructed an AMS consisting of a quadrotor UAV equipped with a 7-DoF Franka Panda manipulator. The UAV control space is defined as position changes in Cartesian coordinates and Yaw rotation; the manipulator control space includes 7 joint poses and the gripper state. This configuration constitutes a high-dimensional control problem characterized by high redundancy and coupling.

\subsection{Dataset Construction}
Data collection utilizes the Isaac Sim-based environment, emphasizing diversity and physical realism.

\paragraph{Data Collection Strategy.}
To capture atypical aerial coordination patterns, we employ \textbf{Human Teleoperation} rather than scripted generation. Expert demonstrators utilize gamepad interfaces to generate task-oriented trajectories, covering multiple collaborative solutions for the UAV and manipulator within unified tasks.

\paragraph{Multimodal Perception System.}
Tailored to aerial perspectives, the sensor configuration comprises: (1) a UAV front-down RGB-D camera for global bird's-eye views, (2) a manipulator wrist RGB-D camera for local fine-grained observation, and (3) a traditional third-person perspective camera.

\paragraph{Action and State Space.}
The dataset records full proprioceptive information, including the UAV's 4D pose and derivatives (velocity/angular velocity), manipulator joint poses, and end-effector pose. Standardized data interfaces ensure compatibility with the input layers of diverse VLA models.

\paragraph{Scene and Instruction Diversity.}
To enhance generalization, we constructed diverse environments (residential, industrial, outdoor) with varying lighting conditions. Furthermore, LLMs are leveraged to generate natural language instructions featuring complex structures and implicit intents, ensuring dense coverage of the semantic space.

\section{Experiments}

\subsection{VLA Experiments}

We evaluated a series of advanced VLA models to assess their performance in aerial manipulation tasks. While VLA models have demonstrated superior efficacy in fixed-base and ground mobile manipulation tasks, we apply them to AMS for the first time. Our experiments are structured around the following research questions:

\begin{itemize}
    \item[\textbf{Q1:}] Can VLA models maintain superior task completion performance in high-DoF (12-DoF in experiments) robotic system manipulation tasks?
    \item[\textbf{Q2:}] Can VLA models cope with external disturbances in AMS and complete tasks under random base jitter?
    \item[\textbf{Q3:}] What is the dependency of VLA models on visual inputs in AMS tasks, and which viewing angles significantly enhance task completion?
    \item[\textbf{Q4:}] How do VLA models perform in UAV movement and manipulator operation respectively in AMS tasks, and how is their coordination?
    \item[\textbf{Q5:}] Safety constraints are critical in robotic tasks. Do VLA models introduce more severe safety issues in AMS tasks?
\end{itemize}

\subsubsection{Experimental Setup}

\paragraph{Baseline Models.}
To establish a representative benchmark, we evaluate six diverse models: $\boldsymbol{\pi_0}$ \cite{black2026pi0visionlanguageactionflowmodel} and $\boldsymbol{\pi_{0.5}}$ \cite{black2025pi}, Flow Matching-based foundation models pre-trained on cross-embodiment data, represent large-scale transfer capabilities to the aerial domain. \textbf{$\boldsymbol{\pi_{0}\text{-FAST}}$}\cite{pertsch2025fastefficientactiontokenization} serves as an efficiency-optimized variant, utilizing step distillation to enable the high-frequency inference crucial for agile aerial maneuvers. \textbf{ACT} \cite{zhao2023learningfinegrainedbimanualmanipulation} combines CVAEs and Transformers to predict ``Action Chunks,'' effectively modeling long-term dependencies. \textbf{Diffusion Policy} \cite{chi2024diffusionpolicyvisuomotorpolicy} utilizes conditional diffusion for iterative denoising, robustly handling multi-modal distributions essential for floating-base stability.

\paragraph{Dataset \& Training.}
The training dataset is derived from our human-teleoperated simulation data, fully reflecting the diversity of collaborative actions between UAV and manipulator. We adopted a differentiated fine-tuning strategy based on task difficulty: 30 trajectories were used for fine-tuning simple single-step tasks with less physical interaction, while 50 trajectories were used for multi-step long-horizon tasks or tasks with rich physical interactions.

\paragraph{Robustness Evaluation.}
To simulate real flight environments (Sim-to-Real Gap), we introduced dynamic disturbance test scenarios. During testing, Gaussian noise was injected into the UAV's state space to simulate the effects of wind gusts or sensor drift on flight stability. Additionally, we designed a "No Fixed Third-Person View" test, retaining only onboard and wrist cameras, to investigate the model's dependency on global perspectives.

\subsubsection{Main Results and Analysis}
In this section, we report the quantitative performance of baseline models on the four task suites of AIR-VLA. Table \ref{tab:comprehensive_results} presents the comprehensive evaluation results under standard disturbance-free conditions.

\begin{table*}[t]
\centering
\caption{\textbf{Detailed performance evaluation of models across four task suites and overall average.} The table displays normalized scores for sub-metrics and weighted total scores for each model under different task types. Weights are set as $w_{pos}=0.25, w_{arm}=0.25, w_{safe}=0.10, w_{task}=0.40$. Abbreviations: \textbf{Pos}: Base Positioning Accuracy ($S_{pos}$); \textbf{Arm}: Manipulator Efficacy ($S_{arm}$); \textbf{Safe}: Environmental Safety ($S_{safe}$); \textbf{Task}: Task Progression ($S_{task}$); \textbf{Tot}: Weighted Total Score ($S_{total}$).}
\label{tab:comprehensive_results}

\renewcommand{\arraystretch}{1.1}
\setlength{\tabcolsep}{1.5pt}

\resizebox{\textwidth}{!}{%
\begin{tabular}{l ccccc ccccc ccccc ccccc ccccc}
\toprule
\multirow{3}{*}{\textbf{Model}} & \multicolumn{5}{c}{\textbf{Base Manipulation}} & \multicolumn{5}{c}{\textbf{Object \& Spatial}} & \multicolumn{5}{c}{\textbf{Semantic Understanding}} & \multicolumn{5}{c}{\textbf{Long-Horizon}} & \multicolumn{5}{c}{\textbf{Overall Average}} \\
\cmidrule(lr){2-6} \cmidrule(lr){7-11} \cmidrule(lr){12-16} \cmidrule(lr){17-21} \cmidrule(lr){22-26}
 & Pos & Arm & Safe & Task & \textbf{Tot} & Pos & Arm & Safe & Task & \textbf{Tot} & Pos & Arm & Safe & Task & \textbf{Tot} & Pos & Arm & Safe & Task & \textbf{Tot} & Pos & Arm & Safe & Task & \textbf{Tot} \\
\midrule
% -------------------------------------------------------------------------------------------------------
% Placeholder Data
% -------------------------------------------------------------------------------------------------------
ACT 
 & 0.0 & 20.0 & \textbf{100.0} & 0.0 & \textbf{15.0} 
 & 0.0 & 16.0 & \textbf{100.0} & 0.0 & \textbf{14.0} 
 & 0.0 & 16.0 & \textbf{100.0} & 0.0 & \textbf{14.0} 
 & 0.0 & 10.0 & \textbf{100.0} & 0.0 & \textbf{12.5} 
 & 0.0 & 16.5 & \textbf{100.0} & 0.0 & \textbf{13.9} \\

Diffusion Policy
 & 0.0 & 20.0 & \textbf{100.0} & 0.0 & \textbf{15.0} 
 & 0.0 & 16.0 & \textbf{100.0} & 0.0 & \textbf{14.0} 
 & 0.0 & 16.0 & \textbf{100.0} & 0.0 & \textbf{14.0} 
 & 0.0 & 10.0 & \textbf{100.0} & 0.0 & \textbf{12.5} 
 & 0.0 & 16.5 & \textbf{100.0} & 0.0 & \textbf{13.9} \\

$\pi_0\text{-FAST}$
 & 27.5 & 23.9 & 98.1 & 0.0 & \textbf{22.7} 
 & 5.1 & 18.9 & 98.3 & 0.0 & \textbf{15.8} 
 & 6.5 & 16.9 & 99.3 & 0.0 & \textbf{15.8} 
 & 41.0 & 32.6 & 99.4 & 0.0 & \textbf{28.3} 
 & 15.0 & 20.8 & 98.5 & 0.0 & \textbf{18.8} \\

$\pi_{0}$
 & 56.6 & 48.0 & 87.5 & 8.0 & \textbf{38.1} 
 & 58.1 & 38.2 & 75.4 & 2.0 & \textbf{32.4} 
 & 55.7 & 36.1 & 85.1 & 2.00 & \textbf{32.3} 
 & \textbf{65.1} & 38.5 & 85.9 & 2.0 & \textbf{32.3} 
 & 58.8 & 42.0 & 83.5 & 3.50 & \textbf{34.5} \\

$\pi_{0.5}$
 & \textbf{78.8} & \textbf{49.9} & 87.2 & \textbf{11.0} & \textbf{45.3} 
 & \textbf{70.9} & \textbf{44.1} & 81.2 & \textbf{13.5} & \textbf{42.3} 
 & \textbf{65.4} & \textbf{44.7} & 82.7 & \textbf{11.0} & \textbf{40.2} 
 & 62.4 & \textbf{40.8} & 84.8 & \textbf{7.0} & \textbf{37.1} 
 & \textbf{71.3} & \textbf{45.9} & 84.7 & \textbf{10.6} & \textbf{42.0} \\ 
\bottomrule
\end{tabular}%
}
\end{table*}

\paragraph{Overall Performance.}
Experimental results indicate that large-scale pre-trained models, represented by $\pi_{0.5}$ and $\pi_0$, demonstrate significant advantages in the AIR-VLA evaluation, outperforming traditional imitation learning baselines such as ACT and Diffusion Policy across all metrics. However, in the complex domain of aerial manipulation tasks, $\boldsymbol{\pi_0\text{-FAST}}$ exhibits a marked performance decline compared to $\pi_0$, suggesting that the performance degradation induced by distillation and step reduction is significantly amplified in aerial scenarios.

Overall, the results validate the feasibility of transferring VLA paradigms to Aerial Manipulation Systems. Furthermore, they demonstrate that pre-training on massive cross-embodiment data (including mobile robotics) enables models to acquire generalizable mechanical manipulation priors. Notably, even under few-shot fine-tuning settings with only 30--50 demonstrations, foundation models like $\pi_{0.5}$ rapidly adapt to aerial manipulation paradigms unseen during pre-training, achieving the highest performance among all evaluated models. Nevertheless, there remains substantial room for improvement in overall task completion rates. Compared to low-DoF ground-based platforms, the performance of existing VLA models on high-DoF aerial platforms remains suboptimal.

$\pi_0$ achieves its peak success rate in Base Manipulation tasks characterized by minimal environmental interference; however, its performance exhibits a significant decaying trend with the introduction of intense visual distractors, complex semantic instructions, and long-horizon planning requirements. Specifically, within the Object \& Spatial suite, dense visual distractors lead to a marked increase in target misidentification rates. Notably, in spatial understanding tasks, the models exhibit Spatial Grounding Failure: although the correct object category is identified, the agent manipulates an identical object at an incorrect location due to an inability to accurately parse relative positional instructions. In comparison, $\pi_{0.5}$ demonstrates superior robustness against object interference relative to $\pi_0$. In the Semantic Understanding suite, where visual scene complexity is controlled to be comparable to that of the Object \& Spatial suite, we observe a slight performance decline in both $\pi_{0.5}$ and $\pi_0$, suggesting that complex semantic instructions impose a distinct penalty on model performance. In the Long-Horizon suite, we observe that successful executions are predominantly confined to the initial sub-task, with performance on the subsequent sub-task exhibiting notable degradation compared to the first.

Regarding collaborative control, VLA models demonstrate significantly better performance in UAV motion control than in manipulator operation. This is partly attributable to the fact that UAV movement in aerial manipulation is relatively coarse-grained, whereas manipulator operation necessitates extremely high precision. Existing models have successfully acquired UAV navigation strategies to guide the system to feasible operating positions; however, executing fine-grained manipulation at these positions remains a critical technical bottleneck. Furthermore, safety constraints cannot be overlooked. Due to the inherent characteristics of the floating base, collisions and unreasonable physical interactions cause significantly more severe disturbances to the system than in ground-based robotics. Our evaluation revealed destructive interactions with the environment in certain episodes, indicating that addressing safety constraints in Aerial Manipulation Systems remains a pivotal direction for future research.

\paragraph{Robustness Analysis.}
We selected the best-performing model, $\pi_{0.5}$, for UAV dynamic disturbance and third-person lacking tests (results in Table \ref{tab:robustness_test}). Under UAV disturbance, the decline in performance metrics for $\pi_{0.5}$ is relatively small, indicating that VLA models possess a degree of implicit compensation capability. However, in the absence of fixed third-person view input, model performance decays significantly. This reveals the current models' dependency on stable global perspectives; the lack of a global view severely limits the system's executable space and exploration capabilities. Introducing active perception strategies based on onboard UAV cameras may be an effective solution to this problem.

\begin{table}[t]
\centering
\caption{\textbf{Robustness evaluation of $\pi_{0.5}$ under disturbance and perception-deprived conditions.} The table shows absolute scores for each metric, with values in parentheses indicating the performance drop ($\downarrow \Delta$) compared to the Standard condition.}
\label{tab:robustness_test}

\resizebox{\columnwidth}{!}{%
\begin{tabular}{lccccc}
\toprule
\textbf{Condition} & \boldmath{$S_{pos}$} & \boldmath{$S_{arm}$} & \boldmath{$S_{safe}$} & \boldmath{$S_{task}$} & \boldmath{$S_{total}$} \\
\midrule
Standard (None) 
 & \textbf{71.3} & \textbf{45.9} & \textbf{84.7} & \textbf{10.6} & \textbf{42.0} \\ 
\midrule
UAV Disturb.
 & 68.2 \scriptsize{\color{gray}($\downarrow$3.1)} 
 & 45.6 \scriptsize{\color{gray}($\downarrow$0.3)} 
 & 83.2 \scriptsize{\color{gray}($\downarrow$1.5)} 
 & 10.5 \scriptsize{\color{gray}($\downarrow$0.1)} 
 & 41.0 \scriptsize{\color{gray}($\downarrow$1.0)} \\

View Lacking
 & 58.9 \scriptsize{\color{gray}($\downarrow$12.4)} 
 & 35.6 \scriptsize{\color{gray}($\downarrow$10.3)} 
 & 85.6 \scriptsize{\color{gray}($\uparrow$0.9)} 
 & 5.1 \scriptsize{\color{gray}($\downarrow$5.5)} 
 & 34.5 \scriptsize{\color{gray}($\downarrow$7.5)} \\
\bottomrule
\end{tabular}%
}
\end{table}

\subsection{VLM Experiments}
Leveraging the capability of VLMs to integrate visual and linguistic instructions for high-level planning, we conduct a comprehensive evaluation of several mainstream VLMs. Specifically, we assess their planning performance across the dimensions of Process Planning, Spatial Navigation, Object Grounding, and Skill Selection, aiming to verify their potential for aerial manipulation tasks.

\begin{figure*}[t] 
  \vskip 0.2in
  \begin{center}
    \centerline{\includegraphics[width=\textwidth]{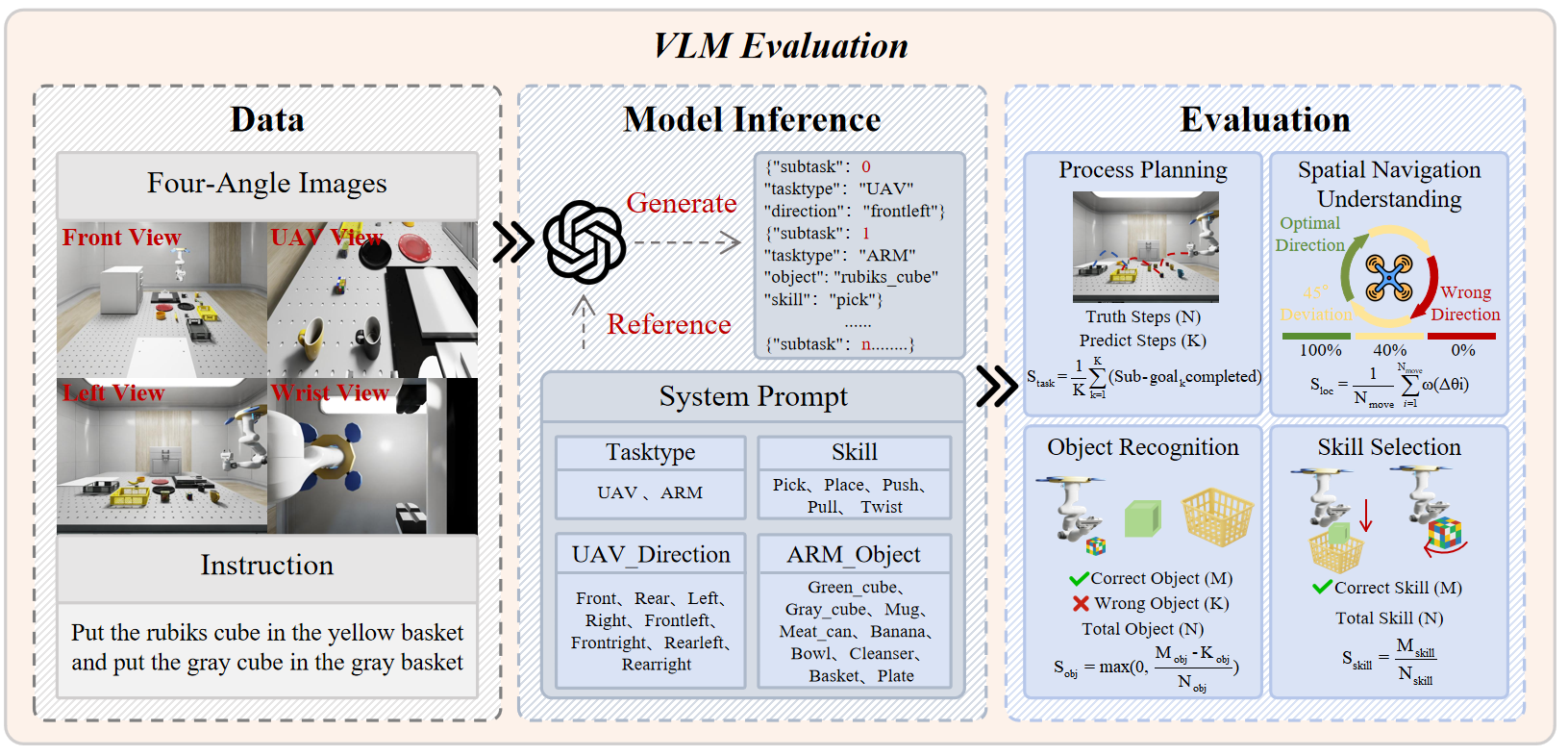}}
    \caption{
      \textbf{Evaluation pipeline for VLM high-level planning capabilities in aerial manipulation tasks.} 
    }
    \label{fig:Evaluation pipeline for VLMs} 
  \end{center}
\end{figure*}

\begin{table*}[t!]
\centering
\caption{\textbf{Detailed evaluation of VLM task planning capabilities.} The table displays normalized sub-metric scores and planning success rates (Succ, \%) for each model across different task scenarios and instruction types. Models are sorted in ascending order by their Overall Average Total score. Abbreviations: \textbf{Exp.}: Explicit; \textbf{Imp.}: Implicit. Weights: $w_{plan}=0.25, w_{loc}=0.25, w_{obj}=0.25, w_{skill}=0.25$.}
\label{tab:vlm_planning_comprehensive}

\renewcommand{\arraystretch}{1.2} % Increase row height slightly
\setlength{\tabcolsep}{1.2pt} % Compress column spacing

\resizebox{\textwidth}{!}{%
\begin{tabular}{ll cccccc cccccc cccccc cccccc}
\toprule
\multirow{3}{*}{\textbf{Model}} & \multirow{3}{*}{\textbf{Instr.}} & \multicolumn{6}{c}{\textbf{Base Manipulation}} & \multicolumn{6}{c}{\textbf{Distractor Interference}} & \multicolumn{6}{c}{\textbf{Long-Horizon}} & \multicolumn{6}{c}{\textbf{Overall Average}} \\
\cmidrule(lr){3-8} \cmidrule(lr){9-14} \cmidrule(lr){15-20} \cmidrule(lr){21-26}
 & & Plan & Loc & Obj & Skill & \textbf{Tot} & Succ & Plan & Loc & Obj & Skill & \textbf{Tot} & Succ & Plan & Loc & Obj & Skill & \textbf{Tot} & Succ & Plan & Loc & Obj & Skill & \textbf{Tot} & Succ \\
\midrule

% ==================== 1. LLaVA-OneVision (Total: 0.37) ====================
\multirow{2}{*}{LLaVA-OV} 
 & Exp. 
 & 10.0 & 38.0 & 0.0 & 80.0 & \textbf{32.0} & 0.0
 & 12.0 & 38.0 & 0.0 & 75.0 & \textbf{31.3} & 0.0
 & 72.0 & 29.3 & 22.5 & 62.5 & \textbf{46.6} & 0.0
 & \multirow{2}{*}{31.8} & \multirow{2}{*}{35.4} & \multirow{2}{*}{9.6} & \multirow{2}{*}{73.2} & \multirow{2}{*}{\textbf{37.5}} & \multirow{2}{*}{0.0}
 \\
 & Imp. 
 & 6.0 & 38.0 & 0.0 & 85.0 & \textbf{32.3} & 0.0
 & 22.0 & 43.0 & 10.0 & 70.0 & \textbf{36.3} & 0.0
 & 68.7 & 26.0 & 25.0 & 66.7 & \textbf{46.6} & 0.0
 &  &  &  &  &  & 
 \\
\midrule

% ==================== 3. Molmo-7B (Total: 0.55) ====================
\multirow{2}{*}{Molmo-7B} 
 & Exp. 
 & 33.0 & 42.0 & 0.0 & 70.0 & \textbf{36.3} & 0.0
 & 38.0 & 47.0 & 0.0 & 50.0 & \textbf{33.8} & 0.0
 & 72.0 & 37.8 & 13.3 & 77.5 & \textbf{50.2} & 0.0
 & \multirow{2}{*}{47.6} & \multirow{2}{*}{40.4} & \multirow{2}{*}{4.9} & \multirow{2}{*}{61.2} & \multirow{2}{*}{\textbf{38.5}} & \multirow{2}{*}{0.0}
 \\
 & Imp. 
 & 27.0 & 37.0 & 0.0 & 55.0 & \textbf{29.8} & 0.0
 & 43.0 & 49.0 & 0.0 & 40.0 & \textbf{33.0} & 0.0
 & 72.4 & 29.5 & 15.8 & 75.0 & \textbf{48.2} & 0.0
 &  &  &  &  &  & 
 \\
\midrule

% ==================== 2. InternVL3.5 (Total: 0.43) ====================
\multirow{2}{*}{InternVL3.5} 
 & Exp. 
 & 23.0 & 0.0 & 30.0 & 100.0 & \textbf{38.3} & 0.0
 & 32.5 & 8.0 & 20.0 & 80.0 & \textbf{35.1} & 0.0
 & 37.7 & 3.5 & 100 & 100 & \textbf{60.3} & 0.0
 & \multirow{2}{*}{31.8} & \multirow{2}{*}{3.2} & \multirow{2}{*}{56.1} & \multirow{2}{*}{89.4} & \multirow{2}{*}{\textbf{45.2}} & \multirow{2}{*}{0.0}
 \\
 & Imp. 
 & 30.0 & 0.0 & 80.0 & 100 & \textbf{52.5} & 0.0
 & 32.0 & 8.0 & 30.0 & 80.0 & \textbf{37.5} & 0.0
 & 35.6 & 0.0 & 76.7 & 76.7 & \textbf{47.2} & 0.0
 &  &  &  &  &  & 
 \\
\midrule

% ==================== 4. GLM-4V (Total: 0.57) ====================
\multirow{2}{*}{GLM-4V} 
 & Exp. 
 & 76.5 & 7.00 & 65.0 & 90.0 & \textbf{59.6} & 10.0
 & 81.0 & 14.0 & 20.0 & 85.0 & \textbf{50.0} & 0.0
 & 76.1 & 26.7 & 36.7 & 75.0 & \textbf{53.6} & 10.0
 & \multirow{2}{*}{79.8} & \multirow{2}{*}{16.8} & \multirow{2}{*}{46.2} & \multirow{2}{*}{84.4} & \multirow{2}{*}{\textbf{56.8}} & \multirow{2}{*}{6.7}
 \\
 & Imp. 
 & 80.0 & 11.0 & 60.0 & 100.0 & \textbf{62.8} & 10.0
 & 79.0 & 13.0 & 40.0 & 80.0 & \textbf{53.0} & 0.0
 & 86.0 & 29.3 & 55.8 & 76.7 & \textbf{62.0} & 10.0
 &  &  &  &  &  & 
 \\
\midrule

% ==================== 6. Qwen2.5-VL (Total: 0.64) ====================
\multirow{2}{*}{Qwen2.5-VL} 
 & Exp. 
 & 94.0 & 37.0 & 75.0 & 100 & \textbf{76.5} & 10.0
 & 92.0 & 46.0 & 50.0 & 90.0 & \textbf{69.5} & 0.0
 & 76.2 & 27.2 & 60.9 & 75.0 & \textbf{59.8} & 20.0
 & \multirow{2}{*}{92.2} & \multirow{2}{*}{38.7} & \multirow{2}{*}{75.1} & \multirow{2}{*}{88.9} & \multirow{2}{*}{\textbf{73.7}} & \multirow{2}{*}{13.3}
 \\
 & Imp. 
 & 100.0 & 37.0 & 100.0 & 100.0 & \textbf{84.3} & 20.0
 & 98.0 & 44.0 & 95.0 & 80.0 & \textbf{79.3} & 10.0
 & 93.0 & 41.2 & 70.0 & 88.3 & \textbf{73.1} & 20.0
 &  &  &  &  &  & 
 \\
\midrule

 % ==================== 5. Qwen3-VL (Total: 0.58) ====================
\multirow{2}{*}{Qwen3-VL} 
 & Exp. 
 & 93.0 & 36.0 & 100.0 & 100.0 & \textbf{82.3} & 20.0
 & 100.0 & 49.0 & 90.0 & 100 & \textbf{84.8} & 20.0
 & 98.7 & 43.5 & 90.0 & 90.0 & \textbf{80.5} & 30.0
 & \multirow{2}{*}{96.9} & \multirow{2}{*}{43.5} & \multirow{2}{*}{93.3} & \multirow{2}{*}{96.7} & \multirow{2}{*}{\textbf{82.4}} & \multirow{2}{*}{23.3}
 \\
 & Imp. 
 & 93.0 & 38.0 & 100.0 & 100.0 & \textbf{82.8} & 30.0
 & 100.0 & 44.0 & 90.0 & 100.0 & \textbf{83.5} & 10.0
 & 96.7 & 46.0 & 90.0 & 90.0 & \textbf{80.7} & 30.0
 &  &  &  &  &  & 
 \\

\bottomrule
\end{tabular}%
}
\end{table*}

\subsubsection{Experimental Setup}

\paragraph{Baseline Models.}
To establish a representative benchmark, we analyze six diverse open-source VLMs. \textbf{Molmo-7B-D-0924} \cite{deitke2024molmopixmoopenweights} employs a data-centric strategy via the PixMo dataset for efficient alignment. The Qwen series, comprising \textbf{Qwen3-VL-8B-Instruct} \cite{bai2025qwen3vltechnicalreport} and \textbf{Qwen2.5-VL-7B-Instruct} \cite{bai2025qwen25vltechnicalreport}, excels in adaptability; the former leverages dynamic resolution, while the latter utilizes M-RoPE to unify image and video understanding. \textbf{GLM-4V-9B} \cite{glm2024chatglmfamilylargelanguage} focuses on deep vision-language alignment for high-resolution features. \textbf{InternVL3\_5-8B} \cite{wang2025internvl35advancingopensourcemultimodal} adopts a ``strong vision encoder + strong language decoder'' design, targeting geometric perception. Finally, \textbf{llava-onevision-qwen2-7b-ov-hf} \cite{li2024llavaonevisioneasyvisualtask} balances static and temporal analysis through a unified mixed-data paradigm.

\paragraph{Assessment Protocol.}
In the VLM evaluation phase, we adopted a non-interactive evaluation process. We input the initial four-view images of the task scene and description instructions to the VLM, providing skill library descriptions, output format requirements, and few-shot examples as System Prompts. The VLM is required to generate high-level task planning sequences conforming to specifications, enabling standardized capability assessment.

\subsubsection{Results and Analysis}

Quantitative results demonstrate that \textbf{Qwen3-VL-8B-Instruct} and \textbf{Qwen2.5-VL-7B-Instruct} establish a substantial performance advantage in high-level planning for aerial manipulation tasks. Notably, Qwen3-VL achieves state-of-the-art (SOTA) performance across all baseline models in four core dimensions: Process Planning, Spatial Navigation, Object Grounding, and Skill Selection, highlighting its exceptional potential as a multimodal ``brain.''

Regarding metric distribution, the capability profiles across models exhibit high consistency. While demonstrating strong performance in semantic understanding and skill matching, all models reveal a significant deficiency in \textbf{Spatial Navigation}. In-depth analysis indicates that this lack of 3D spatial awareness is the primary bottleneck limiting the end-to-end planning Success Rate.

In terms of instruction robustness, the performance gap between explicit, structured instructions and implicit, complex instructions is negligible. This suggests that current VLMs possess powerful semantic generalization capabilities, effectively aligning unstructured linguistic inputs with specific operational intents. Regarding task scenarios, VLMs demonstrate excellent temporal stability; their performance in Long-Horizon tasks does not exhibit significant deterioration, standing in sharp contrast to the \textbf{performance decay} often observed in traditional VLA models over long sequences. However, in scenarios introducing visual Distractor Interference, most models show a slight decline in object grounding metrics, reflecting the challenge complex backgrounds pose to fine-grained visual perception.

In summary, VLMs hold immense potential for high-level planning in aerial manipulation, particularly in mitigating the long-horizon reasoning limitations of VLA models. However, overcoming the critical bottleneck of \textbf{spatial orientation understanding} remains a pivotal challenge for achieving highly reliable aerial agents in the future.

\section{Conclusion}
This paper presents \textbf{AIR-VLA}, the first full-stack VLA benchmark for Aerial Manipulation Systems. Through the construction of a simulation environment, a physics-aware dataset, and a multi-dimensional evaluation framework, we systematically quantify the capability boundaries of current mainstream VLA models in 3D aerial manipulation tasks. Our findings reveal that while transferring pre-trained VLA models to aerial platforms is feasible, existing models still face severe challenges in handling floating-base dynamic coupling, 3D long-horizon planning, and disturbance rejection. AIR-VLA not only provides a standardized evaluation tool for the community but also highlights a critical future direction: tightly integrating explicit multi-body dynamics priors with the generalized reasoning capabilities of large models. We hope this benchmark will inspire further exploration into general-purpose aerial embodied intelligence, propelling robotics from the 2D ground into the vast 3D space.

\clearpage

\section*{Impact Statement}
This paper presents work whose goal is to advance the field of aerial robotics and embodied AI. The primary societal implication of this research is the potential development of more autonomous and versatile unmanned aerial vehicles (UAVs), which could benefit applications such as search and rescue, logistics, and infrastructure inspection. However, integrating VLMs into physical control systems introduces challenges related to safety, reliability, and the potential for unintended physical actions due to model hallucinations or grounding failures. By systematically evaluating these failure modes, our work aims to promote the safe and robust deployment of such systems. We do not feel there are other specific negative societal consequences that must be highlighted here.

\bibliography{airvla}
\bibliographystyle{icml2026}

\newpage
\appendix
\onecolumn

\section{Benchmark Implementation Details}
\label{app:benchmark_impl}

\subsection{Task Description and Taxonomy}

\textbf{Task Overview.} AIR-VLA aims to establish a comprehensive evaluation platform covering the full skill spectrum of aerial mobile manipulation. We have constructed a diverse family of tasks comprising 60 atomic and composite tasks, logically categorized into four core suites: Base Manipulation, Object \& Spatial, Semantic Understanding, and Long-Horizon. The complete task list is detailed in Table \ref{tab:full_task_list}. These tasks range from omnidirectional pick-and-place in 3D space and precision insertion in confined spaces, to articulated object interaction involving drawers and cabinets, and twist-and-rotate operations requiring torque control. Furthermore, the benchmark integrates hybrid navigation-manipulation tasks, requiring agents to seamlessly switch between extensive scene movement and local fine-grained manipulation. Beyond basic motor skills, this benchmark emphasizes deep evaluation of advanced embodied intelligence across multiple dimensions, including voxel-level spatial perception, dynamic pose compensation of the floating base, cross-modal semantic generalization, and long-horizon temporal logical planning.

\begin{figure}[ht]
  \vskip 0.2in
  \begin{center}
    \centerline{\includegraphics[width=\columnwidth]{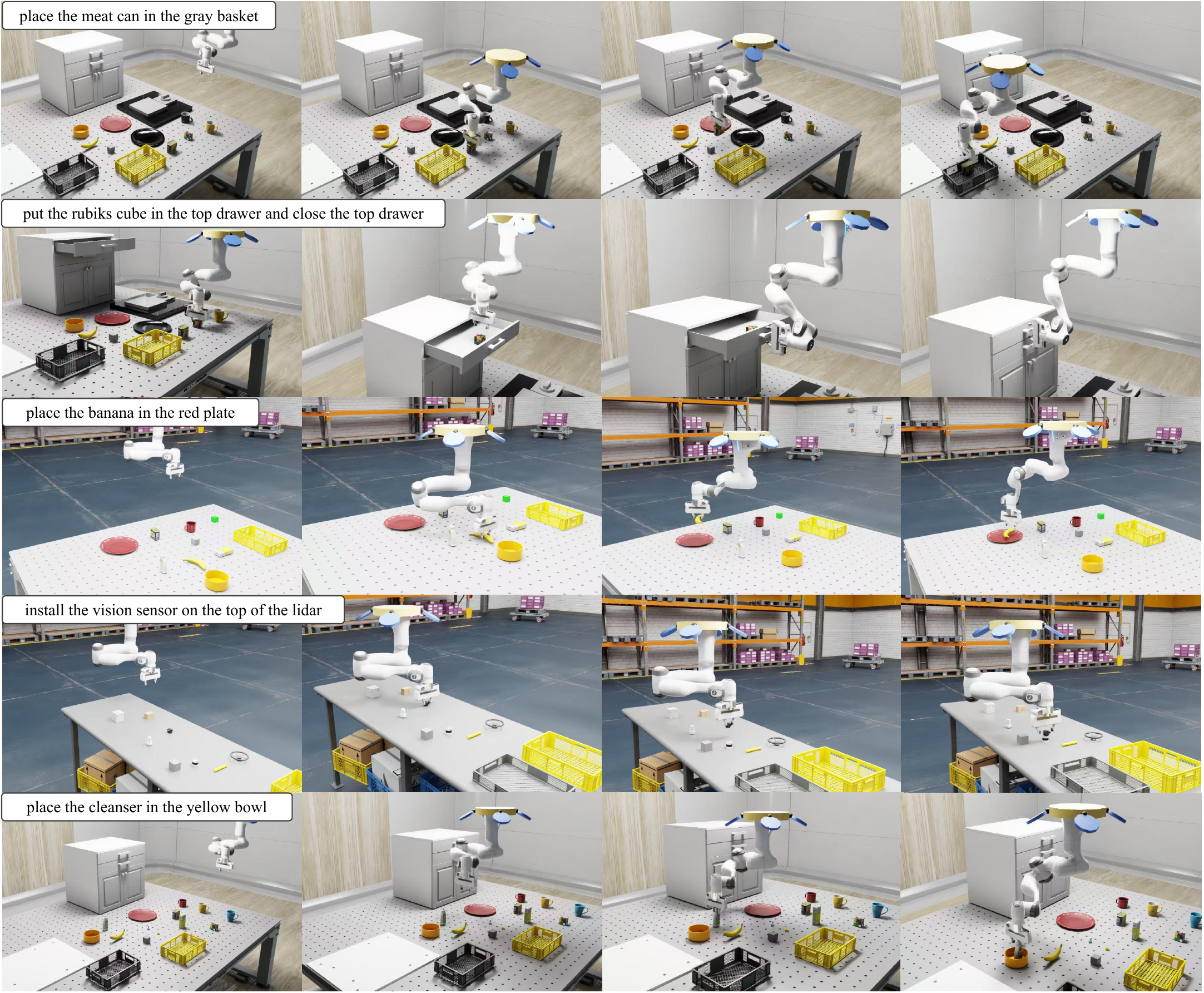}}
    \caption{
      Task examples in AIR-VLA dataset. 
    }
    \label{icml-historical}
  \end{center}
\end{figure}

\textbf{Coupled Dynamics and Long-Horizon Reasoning.} Unlike traditional benchmarks that often treat navigation and manipulation as decoupled phases or focus solely on 2D planar movement, the core feature of AIR-VLA is the emphasis on spatio-temporal coupling between the UAV base and the manipulator. In our context, "reasoning" is endowed with a dual meaning: first, kinematic feasibility reasoning, where the agent must real-time assess whether the target is within the manipulator's reachable workspace under the floating state of the UAV with drift and jitter; second, long-horizon temporal planning, where early interaction actions (e.g., opening a drawer and maintaining a specific opening) must create necessary preconditions for subsequent geometric constraints (e.g., placing an object inside). Our dataset statistics indicate that the average episode length is approximately 475 time steps, with the Long-Horizon suite exceeding 600 time steps (at 20Hz control frequency), significantly surpassing traditional tabletop manipulation datasets. This long-horizon characteristic requires policies not only to maintain logical consistency over extended periods but also to continuously output control signals to stabilize the base and resist environmental disturbances while executing fine-grained sub-tasks.

\subsection{Observation Space}

To support perception and decision-making in complex environments, AIR-VLA constructs a multimodal observation space rich in semantic information. In terms of visual perception, the system synchronously captures RGB-D image streams from four perspectives: a front-down chassis camera provides a first-person flight view for global path planning and target approach; an eye-in-hand camera mounted on the manipulator end-effector resolves occlusion issues and supports fine manipulation; while global overhead and side-view cameras provide third-person god views to assist state estimation and training supervision. Regarding proprioception, the system records full state vectors at high frequency, including the UAV's position, orientation quaternion, linear velocity, and angular velocity in the world frame, the manipulator's 7 joint poses and velocities, as well as the end-effector's 6-DoF pose and gripper state. All sensory data is strictly timestamp-synchronized, providing a precise data alignment foundation for multimodal learning.

\begin{figure}[ht]
  \vskip 0.2in
  \begin{center}
    \centerline{\includegraphics[width=\columnwidth]{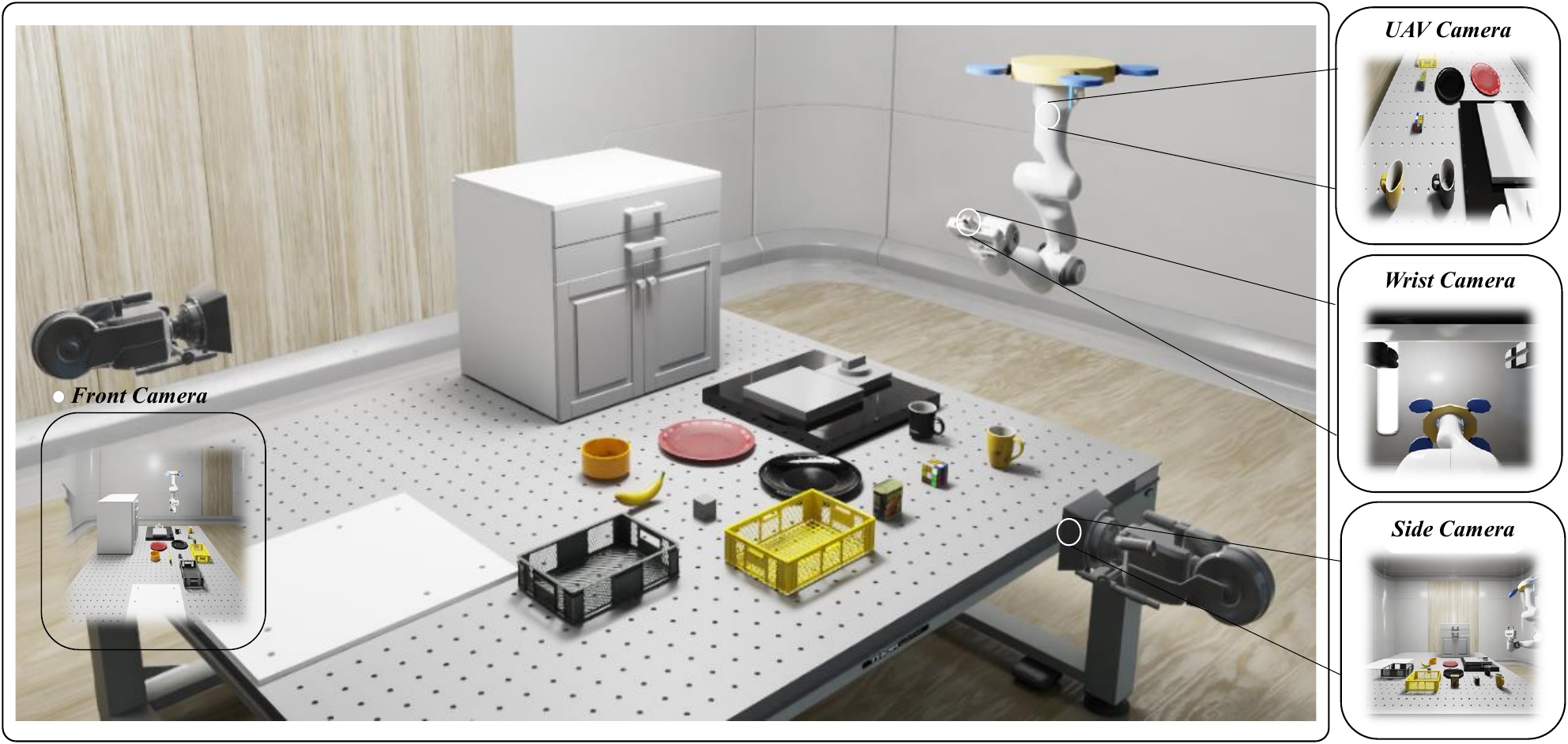}}
    \caption{
      Observation Space. This figure illustrates instances of observations from various camera perspectives.
    }
    \label{icml-historical}
  \end{center}
\end{figure}

\subsection{Evaluation Metrics}
\label{sec:appendix_metrics}

\subsubsection{VLA Evaluation Metrics}

Traditional binary metrics fail to capture continuous state changes in aerial manipulation. To evaluate UAV movement, coordination, and safety at a fine granularity, we propose a four-dimensional system. The total score $S_{total}$ is a weighted sum of metrics $S_i \in [0, 100]$, where the maximum possible score is 100:
\begin{equation}
    S_{total} = w_{pos} S_{pos} + w_{arm} S_{arm} + w_{safe} S_{safe} + w_{task} S_{task}
\end{equation}

\paragraph{Base Positioning Accuracy ($S_{pos}$)}
This metric assesses if the UAV reaches a Feasible Workspace (collision-free Inverse Kinematics (IK) solution). We model positioning quality relative to optimal distance $\mu$ using a Gaussian decay function, scored hierarchically by interaction state on a 0-100 scale:
\begin{equation}
    s_{pos}^{(i)} = 
    \begin{cases}
        100, & \text{if operation succeeds} \\
        100 \cdot \mathcal{G}(d_i; \lambda_{act}), & \text{if interaction attempted} \\
        100 \cdot \mathcal{G}(d_i; \lambda_{hov}), & \text{if hovering only}
    \end{cases}
\end{equation}
where $\mathcal{G}(d; \lambda) = \lambda \cdot \exp\left(-\frac{(d - \mu)^2}{2\sigma^2}\right) \cdot \mathbb{I}(d_{min} \le d \le d_{max})$. $\lambda_{act} > \lambda_{hov}$ rewards active attempts.

\paragraph{Manipulator Efficacy ($S_{arm}$)}
To quantify resilience against floating-base jitter, we combine orientation consistency $S_{ori} \in \{0, 1\}$ (Euler angle alignment) and interaction precision $s_{int}^{(i)}$ to form a score out of 100:
\begin{equation}
    S_{arm} = 100 \cdot \left( \beta S_{ori} + \frac{1-\beta}{N}\sum_{i=1}^{N} s_{int}^{(i)} \right)
\end{equation}
where interaction precision uses a Gaussian kernel on end-effector distance $d_{ee}$:
\begin{equation}
    s_{int}^{(i)} = \exp\left(-\frac{d_{ee}^2}{2\sigma_{arm}^2}\right) \cdot \mathbb{I}(d_{ee} \le d_{thresh})
\end{equation}

\paragraph{Environmental Disturbance Safety ($S_{safe}$)}
Collision risk is measured via the total displacement $\Delta_{total}$ of non-target objects. The score decays from 100 (static environment) to 0 (exceeding limit $\Delta_{limit}$):
\begin{equation}
    S_{safe} = 100 \cdot \exp\left(-\frac{\Delta_{total}^2}{2\sigma_{safe}^2}\right) \cdot \mathbb{I}(\Delta_{total} < \Delta_{limit})
\end{equation}

\paragraph{Task Progression ($S_{task}$)}
For Long-Horizon Tasks with $K$ sub-goals, we measure the percentage of completion to reveal planning decay:
\begin{equation}
    S_{task} = \frac{100}{K} \sum_{k=1}^{K} \mathbb{I}(\text{Sub-goal}_k \text{ completed})
\end{equation}

\subsubsection{VLM Task Planning Evaluation Metrics}

We evaluate high-level planning across process logic, spatial understanding, object grounding, and skill selection using scores scaled to [0, 100]. The total weighted score is:
\begin{equation}
    S_{total} = w_{plan} S_{plan} + w_{loc} S_{loc} + w_{obj} S_{obj} + w_{skill} S_{skill}
\end{equation}

\paragraph{Process Planning Capability ($S_{plan}$)}
Evaluates action sequence correctness via \textbf{Type Accuracy} (agent match $M$) and \textbf{Length Consistency} (step count $K$ vs. ground truth $N$), scaled to 100:
\begin{equation}
    S_{plan} = 100 \cdot \left[ \alpha \cdot \frac{M}{N} + (1-\alpha) \cdot \max\left(0, 1 - \frac{|K - N|}{N}\right) \right]
\end{equation}

\paragraph{Spatial Navigation Understanding ($S_{loc}$)}
Measures UAV movement logic. For $N_{move}$ steps, we score the deviation $\Delta \theta_i$ between predicted and optimal directions using a discrete point system:
\begin{equation}
    \omega(\Delta \theta_i) = 
    \begin{cases}
        100, & \text{if } \Delta \theta_i = 0^\circ \text{ (Correct)} \\
        40, & \text{if } \Delta \theta_i = 45^\circ \text{ (Neighboring)} \\
        0, & \text{otherwise}
    \end{cases}
\end{equation}
\begin{equation}
    S_{loc} = \frac{1}{N_{move}} \sum_{i=1}^{N_{move}} \omega(\Delta \theta_i)
\end{equation}

\paragraph{Object Grounding Capability ($S_{obj}$)}
Penalizes hallucinations during manipulation. With $M_{obj}$ correct and $K_{obj}$ incorrect predictions out of $N_{obj}$ total objects, the score is calculated out of 100:
\begin{equation}
    S_{obj} = 100 \cdot \max\left(0, \frac{M_{obj} - K_{obj}}{N_{obj}}\right)
\end{equation}

\paragraph{Skill Selection Capability ($S_{skill}$)}
The coverage percentage of correctly selected atomic skills $M_{skill}$ out of $N_{skill}$ total skills:
\begin{equation}
    S_{skill} = 100 \cdot \frac{M_{skill}}{N_{skill}}
\end{equation}

\section{Dataset Construction}
\label{app:dataset_construction}

\subsection{Scene Diversity}

To minimize the Sim-to-Real gap and provide diverse visual inputs, we built a simulation environment based on NVIDIA Isaac Sim. We designed six variant environments covering three core space types: household, industrial, and outdoor settings.

\subsection{Assets Library}

Our asset library leverages the Universal Scene Description (USD) ecosystem of Isaac Sim and deeply integrates high-quality assets from ShapeNet and Google Scanned Objects. The library contains objects of various categories, covering a wide range of interactive objects from rigid bodies to articulated mechanisms. To ensure the realism of contact dynamics, all objects undergo rigorous physical property calibration, including convex decomposition optimization of collision meshes, inertia tensor calculation for mass distribution, and fine-tuning of surface friction coefficients.

\begin{table}[H]
  \caption{Statistics of interactive assets in the simulation environment.}
  \label{tab:asset-list}
  \begin{center}
    \begin{small}
      \begin{sc}
        \begin{tabular}{lc}
          \toprule
          Asset Category & Quantity \\
          \midrule
          Stove           & 1 \\
          Mug             & 4 \\
          Cabinet         & 1 \\
          Toy             & 2 \\
          Snack           & 3 \\
          Ingredient      & 1 \\
          Condiment       & 2 \\
          Bowl            & 3 \\
          Box             & 5 \\
          Box container   & 4 \\
          Tray            & 3 \\
          Household       & 2 \\
          Industrial part & 7 \\
          Table           & 3 \\
          \bottomrule
        \end{tabular}
      \end{sc}
    \end{small}
  \end{center}
  \vskip -0.1in
\end{table}

\subsection{Data Collection Pipeline}

To capture natural and diverse action distributions and avoid the behavioral rigidity of rule-based scripts, we employed a "Human-in-the-Loop" simulation teleoperation data collection scheme. Operators synergistically control the UAV's flight and the manipulator's fine movements via gamepad or keyboard. Successful task segments are retained and processed into formats including raw Isaac Sim data, Robomimic-compatible HDF5 format, and LeRobot standardized format adapted for the Hugging Face open-source ecosystem, satisfying the requirements of different training frameworks.

\subsection{VLM Evaluation Setup}

To quantitatively evaluate the task planning capabilities of VLMs, we defined a standardized task description format based on JSON. This format structurally articulates the ideal execution flow of long-horizon tasks, detailing sub-task decomposition logic, execution agents (UAV or ARM), target objects, and required skills. This structured data serves as Ground Truth for automatically evaluating the correctness of VLM output planning logic.

\begin{lstlisting}[language=json, basicstyle=\small\ttfamily, breaklines=true]
[
  {
    "subtask": 0,
    "tasktype": "UAV",
    "direction": "frontleft"
  },
  {
    "subtask": 1,
    "tasktype": "ARM",
    "object": "rubiks_cube",
    "skill": "pick"
  }
]
\end{lstlisting}

\subsection{Instruction Generation and Augmentation}

To comprehensively evaluate the generalization capabilities of models in natural language understanding, we developed an instruction augmentation engine leveraging LLMs. For each specific task, the LLM is prompted to generate novel instructions that retain complete action semantics while employing diverse linguistic variations, including object rephrasing, implicit action inference, and environmental descriptions. This approach ensures the inclusion of unseen yet semantically equivalent instructions within the training set, enabling effective assessment of the model's semantic robustness against unstructured linguistic inputs.The specific prompt utilized for instruction generation is presented below:

\begin{lstlisting}[
    basicstyle=\ttfamily\small, % 打字机字体
    breaklines=true,            % 自动换行
    frame=single,               % 单线边框
    caption={Prompt for instruction generation.},
    label={lst:prompt}
]
The current action description is a bit one-dimensional. I need to make some modifications: 1. Replace the description of the object, 2. Replace the description of the action, 3. Add additional semantic scene information, 4. It can be expressed in a more conversational way.
\end{lstlisting}

\section{Experimental Implementation}
\label{sec:exp_impl}

To establish robust baselines and evaluate the transferability of VLA models to the aerial manipulation domain, we selected five representative architectures for full-parameter fine-tuning: $\boldsymbol{\pi_0}$, $\boldsymbol{\pi_{0.5}}$, $\boldsymbol{\pi_{0}\text{-FAST}}$, \textbf{ACT} (Action Chunking Transformer), and \textbf{Diffusion Policy}.

\paragraph{Training Configuration.}
All models accept RGB images from four perspectives (resized to $224 \times 224$) alongside 7-DoF manipulator joint positions and gripper states as inputs. The models predict a hybrid action vector comprising 4-DoF relative UAV displacement commands, 7-DoF manipulator joint position targets, and the gripper state. We utilized default hyperparameters for all models. Using training volume as the standardization metric, all models were trained for 10,000 epochs with identical batch sizes. All experiments were executed in parallel on two NVIDIA A100 GPUs (80GB VRAM). We utilized the official open-source implementations for each model while maintaining default architectural hyperparameters. This setup was chosen to verify the potential of general-purpose VLA architectures to adapt to high-dimensional aerial manipulation tasks without domain-specific modifications.

\paragraph{Evaluation Protocol.}
During the evaluation phase, we maintained consistency with the training data regarding the formatting of images, language instructions, observation states, and action states, adopting a control frequency of 20Hz. For the inference strategy, specifically for $\pi_0$, $\pi_{0.5}$ and $\pi_{0}\text{-FAST}$, we executed the first 5 actions of the predicted action chunk at each inference step. For Diffusion Policy and ACT, we directly executed the predicted actions. For each specific task, we conducted 10 independent evaluation trials and reported the average metrics across these trials as the final result.

\section{Full Task List}

\begin{small}
\begin{longtable}{p{4.5cm} p{3.0cm} p{7.5cm}}
\caption{Detailed Task List of AIR-VLA (Summary Version)} \label{tab:full_task_list} \\

\toprule
\textbf{Task Name} & \textbf{Suite} & \textbf{Description} \\
\midrule
\endfirsthead

\multicolumn{3}{c}{{\bfseries \tablename\ \thetable{} -- Continued}} \\
\toprule
\textbf{Task Name} & \textbf{Suite} & \textbf{Description} \\
\midrule
\endhead

\midrule
\multicolumn{3}{r}{{Continued on next page...}} \\
\endfoot

\bottomrule
\endlastfoot

% --- Content Starts Here ---

Standard Pick \& Place & Base Manip. & Basic task: Pick gray cube, place in yellow basket. \\
Place on Flat Surface & Base Manip. & Precision placement: Pick cube, place stably on plate. \\
Color Distractor (Green) & Base Manip. & Color discrimination: Pick green cube amidst gray distractors, place in basket. \\
Container Geometry (Bowl) & Base Manip. & Container adaptation: Place green cube into small-aperture bowl. \\
Deformable Object (Banana) & Base Manip. & Irregular object: Pick banana and place in basket, handling non-rigid grasping. \\
Irregular on Plate & Base Manip. & Placement stability: Place banana stably on plate. \\
Tool Manipulation (Mug) & Base Manip. & Handled object: Pick red mug and place on plate. \\
Precise Insertion (Mug) & Base Manip. & Fine manipulation: Insert mug accurately into bowl, avoiding collision. \\
Cylinder Manipulation & Base Manip. & Pose constraint: Pick meat can and place upright on plate. \\
Tall Object (Cleanser) & Base Manip. & Center of mass control: Pick tall cleanser bottle and place in deep basket. \\
Variant: Banana in Box & Base Manip. & Variant: Place cube in purple storage box. \\
Variant: Teddy Bear & Base Manip. & Complex model: Pick teddy bear and place in purple storage box. \\
Variant: Meat Can in Box & Base Manip. & Variant: Place meat can in purple storage box. \\
Variant: Chocolate Box & Base Manip. & Box-shaped object: Pick chocolate box and place in purple storage box. \\
Variant: Wood Block & Base Manip. & Textured object: Pick wood block and place in purple storage box. \\
Variant: Tomato Can & Base Manip. & Variant: Place chocolate box in red bowl. \\
Variant: Teddy in Bowl & Base Manip. & Size adaptation: Place large teddy bear in bowl. \\
Variant: Banana in Bowl & Base Manip. & Variant: Adapt banana curvature to bowl shape. \\
Stacking: Can on Wood & Base Manip. & Simple stacking: Place meat can on top of wood block. \\
Stacking: Box on Wood & Base Manip. & Simple stacking: Place chocolate box on top of wood block. \\
Target Color (Yellow) & Object \& Spatial & Multi-container grounding: Identify yellow basket among multiple and place cube. \\
Target Color (Gray) & Object \& Spatial & Low-contrast grounding: Identify gray basket similar to ground color and place. \\
Target Category (Banana) & Object \& Spatial & Category \& location: Identify banana and place in specified gray basket. \\
Pick Left Cup (Red) & Object \& Spatial & Spatial \& visual matching: Pick the left red cup and place on red plate. \\
Visual Combination (Blue-Red) & Object \& Spatial & Color combination: Place blue cup on red plate, testing instruction following. \\
Cross-Category Transfer & Object \& Spatial & Cross-category move: Place meat can into gray basket. \\
Slim Object in Bowl & Object \& Spatial & Slim object grounding: Place cleanser into yellow bowl. \\
Spatial Ref. (Middle Drawer) & Object \& Spatial & Vertical spatial grounding: Identify middle handle, pull drawer $>0.3$m. \\
Small Object (Bolt) & Object \& Spatial & Tiny object grounding: Identify tiny bolt and place in bowl. \\
Spatial Ref. (Top Drawer) & Object \& Spatial & Dynamic height adaptation: Ascend to identify top handle, pull drawer $>0.3$m. \\
Pick Left Box (White) & Object \& Spatial & Spatial constraint: Pick the white box on the left and place in the left basket. \\
Variant: Green Nut & Object \& Spatial & Small part: Pick green nut and place in yellow basket. \\
Assembly (Camera on Lidar) & Object \& Spatial & Fine assembly: Place camera module on top of LiDAR. \\
Non-Prehensile (Push Cube) & Object \& Spatial & Non-prehensile: Push cube next to gray basket. \\
Clear Desk (Push Box) & Object \& Spatial & Clearing action: Push brown box off the table. \\
Synonym: Container & Semantic Und. & Hypernym: "Container" $\to$ Basket. Action: Cube $\to$ Yellow Basket \\
Synonym: Block \& Bin & Semantic Und. & Double synonym: "Block" $\to$ Cube, "Bin" $\to$ Basket. \\
Verb: Drop & Semantic Und. & Verb nuance: "Drop" implies lower precision release. Action: Banana $\to$ Gray Basket \\
Synonym: Cup \& Dish & Semantic Und. & Noun substitution: "Cup" $\to$ Mug, "Dish" $\to$ Plate. \\
Preposition: Onto & Semantic Und. & Preposition: "Onto" emphasizes surface support. Action: Blue Mug $\to$ Red Plate \\
Material: Tin & Semantic Und. & Material reference: "Tin" $\to$ Metal Can. Action: Meat Can $\to$ Gray Basket \\
Function: Detergent & Semantic Und. & Function reference: "Detergent" $\to$ Cleanser. Action: Cleanser $\to$ Yellow Bowl \\
Synonym: Center & Semantic Und. & Position semantic: "Center" $\to$ Middle Drawer. Pull action. \\
Attribute: Metal & Semantic Und. & Attribute reference: "Metal" $\to$ Bolt. Place in bowl. \\
Superlative: Highest & Semantic Und. & Extreme semantic: "Highest" $\to$ Top Drawer. Pull action. \\
Negation Constraint & Semantic Und. & Negation: "Not the brown one...". Exclude brown, pick white box. \\
Contextual: Spare Part & Semantic Und. & Contextual: "Spare part" refers to green nut, place in parts box. \\
Jargon: Vision Sensor & Semantic Und. & Jargon: "Vision sensor" $\to$ Camera. Place on LiDAR. \\
Preposition: By / Don't & Semantic Und. & Position \& Negation: "By" + "Don't put in". Push action. \\
Abstract Intent: Clear & Semantic Und. & Abstract intent: "Clear the desk" $\to$ Push object off table. \\
Multi-Object Sorting & Long-Horizon & Multi-step sorting: 1. Rubiks $\to$ Yellow; 2. Cube $\to$ Gray. \\
Batch Collection & Long-Horizon & Batch collection: 1. Rubiks $\to$ Yellow; 2. Cube $\to$ Yellow. \\
Color-Matched Sorting & Long-Horizon & Color matching: 1. Black Mug $\to$ Black Plate; 2. Yellow Mug $\to$ Red Plate. \\
Cross-Color Sorting & Long-Horizon & Spatial crossing: 1. Black Mug $\to$ Left(Red); 2. Yellow Mug $\to$ Right(Black). \\
Category Sorting & Long-Horizon & Category sorting: 1. Food(Can) $\to$ Right; 2. Fruit(Banana) $\to$ Right. \\
Stacking \& Nesting & Long-Horizon & Nesting dependency: 1. Move Bowl $\to$ Plate; 2. Place Cube $\to$ Inside Bowl. \\
Tower Stacking & Long-Horizon & Stacking balance: 1. Place Rubiks $\to$ Basket; 2. Place Cube $\to$ On Rubiks. \\
Hybrid Interaction (Stove) & Long-Horizon & Hybrid skill: 1. Rotate Switch $>75^{\circ}$; 2. Place Bowl $\to$ Stove. \\
Drawer Interaction & Long-Horizon & Mechanism interaction: 1. Pull Top Drawer; 2. Place Banana $\to$ Inside. \\
Store \& Close & Long-Horizon & Full storage: 1. Place Rubiks $\to$ Drawer; 2. Close Drawer. \\

\end{longtable}
\end{small}

\end{document}